\documentclass[10pt,twocolumn,letterpaper]{article}

\usepackage{cvpr}              %

\definecolor{cvprblue}{rgb}{0.21,0.49,0.74}
\usepackage[pagebackref,breaklinks,colorlinks,allcolors=cvprblue]{hyperref}

\usepackage[utf8]{inputenc} %
\usepackage[T1]{fontenc}    %
\usepackage{hyperref}       %
\usepackage{url}            %
\usepackage{booktabs}       %
\usepackage{amsfonts}       %
\usepackage{nicefrac}       %
\usepackage{microtype}      %
\usepackage{xcolor}         %
\usepackage{mathtools}
\usepackage{amssymb}
\usepackage{algorithm}
\usepackage{algorithmicx}
\usepackage{algpseudocode}   
\usepackage{tikz}           %
\usepackage{tikz-cd}        %
\usepackage{courier}        %
\usepackage{bm}             %
\usepackage{physics}        %
\usepackage{mdframed}       %
\usepackage{multirow}       %
\usepackage{wrapfig}        %
\usepackage{todonotes}      %
\usepackage{dsfont}         %
\usepackage{CJK}            %
\usepackage{array}          %
\usepackage{makecell}       %
\usepackage{slashed}        %
\usepackage{dashbox}        %
\usepackage{colortbl}       %
\usepackage{amsthm}         %
\usepackage[many]{tcolorbox}%
\newcommand{\revise}{\textsc{ReViSe}\xspace}

\newcommand{\limautoref}[1]{\hyperref[lim:#1]{\textbf{\textcolor{firebrick}{(#1)}}}}
\newcommand{\bea}{\begin{eqnarray}}
\newcommand{\eea}{\end{eqnarray}}

\definecolor{DarkGreen}{rgb}{0.0, 0.5, 0.0}
\definecolor{lightyellow}{rgb}{1.0, 0.95, 0.7}
\definecolor{Blue}{rgb}{0, 0, 0.8}
\definecolor{darkgreen}{rgb}{0,0.40,0}
\definecolor{firebrick}{rgb}{0.698,0.133,0.133}

\definecolor{colorA}{rgb}{1,0,0}
\definecolor{colorB}{rgb}{0,0.3,1}
\definecolor{colorC}{rgb}{0.9,0.8,0.2}
\definecolor{colorD}{rgb}{0,0.65,0}
\definecolor{lesslightgray}{rgb}{0.5,0.5,0.5}
\definecolor{light-gray}{gray}{0.95}

\let\cite\citep %
\usepackage{amsmath}
\makeatletter
\def\th@remark{%
   \thm@headfont{\bfseries}%
   \normalfont %
   \thm@preskip\topsep \divide\thm@preskip\tw@
   \thm@postskip\thm@preskip
}
\makeatother
\theoremstyle{definition}

\tcolorboxenvironment{theorem}{
  breakable,
  colback=black!10,
  colframe=white,%
  width=\dimexpr\linewidth+10pt\relax,%
  enlarge left by=-5pt,%
  enlarge right by=-5pt,%
  boxsep=5pt,%
  boxrule=0pt,
  left=0pt,right=0pt,top=0pt,bottom=0pt,
  sharp corners,
  before skip=\topsep,
  after skip=\topsep
}

\tcolorboxenvironment{lemma}{
  breakable,
  colback=black!10,
  colframe=white,%
  width=\dimexpr\linewidth+10pt\relax,%
  enlarge left by=-5pt,%
  enlarge right by=-5pt,%
  boxsep=5pt,%
  boxrule=0pt,
  left=0pt,right=0pt,top=0pt,bottom=0pt,
  sharp corners,
  before skip=\topsep,
  after skip=\topsep
}

\tcolorboxenvironment{corollary}{
  breakable,
  colback=black!10,
  colframe=white,%
  width=\dimexpr\linewidth+10pt\relax,%
  enlarge left by=-5pt,%
  enlarge right by=-5pt,%
  boxsep=5pt,%
  boxrule=0pt,
  left=0pt,right=0pt,top=0pt,bottom=0pt,
  sharp corners,
  before skip=\topsep,
  after skip=\topsep
}

\theoremstyle{definition}

\tcolorboxenvironment{definition}{
  breakable,
  colback=black!10,
  colframe=white,%
  width=\dimexpr\linewidth+10pt\relax,%
  enlarge left by=-5pt,%
  enlarge right by=-5pt,%
  boxsep=5pt,%
  boxrule=0pt,
  left=0pt,right=0pt,top=0pt,bottom=0pt,
  sharp corners,
  before skip=\topsep,
  after skip=\topsep
}
\theoremstyle{remark}

\tcolorboxenvironment{assumption}{
  breakable,
  colback=black!10,
  colframe=white,%
  width=\dimexpr\linewidth+10pt\relax,%
  enlarge left by=-5pt,%
  enlarge right by=-5pt,%
  boxsep=5pt,%
  boxrule=0pt,
  left=0pt,right=0pt,top=0pt,bottom=0pt,
  sharp corners,
  before skip=\topsep,
  after skip=\topsep
}

\tcolorboxenvironment{problem}{
  breakable,
  colback=black!10,
  colframe=white,%
  width=\dimexpr\linewidth+10pt\relax,%
  enlarge left by=-5pt,%
  enlarge right by=-5pt,%
  boxsep=5pt,%
  boxrule=0pt,
  left=0pt,right=0pt,top=0pt,bottom=0pt,
  sharp corners,
  before skip=\topsep,
  after skip=\topsep
}

\numberwithin{equation}{section} \numberwithin{theorem}{section}
\numberwithin{proposition}{section} \numberwithin{definition}{section}
\numberwithin{lemma}{section} \numberwithin{assumption}{section}
\numberwithin{remark}{section}

\makeatletter
\let\save@mathaccent\mathaccent
\newcommand*\if@single[3]{%
    \setbox0\hbox{${\mathaccent"0362{#1}}^H$}%
    \setbox2\hbox{${\mathaccent"0362{\kern0pt#1}}^H$}%
    \ifdim\ht0=\ht2 #3\else #2\fi
}
\newcommand*\rel@kern[1]{\kern#1\dimexpr\macc@kerna}
\newcommand*\widebar[1]{\@ifnextchar^{{\wide@bar{#1}{0}}}{\wide@bar{#1}{1}}}
\newcommand*\wide@bar[2]{\if@single{#1}{\wide@bar@{#1}{#2}{1}}{\wide@bar@{#1}{#2}{2}}}
\newcommand*\wide@bar@[3]{%
    \begingroup
    \def\mathaccent##1##2{%
        \let\mathaccent\save@mathaccent
        \if#32 \let\macc@nucleus\first@char \fi
        \setbox\z@\hbox{$\macc@style{\macc@nucleus}_{}$}%
        \setbox\tw@\hbox{$\macc@style{\macc@nucleus}{}_{}$}%
        \dimen@\wd\tw@ \advance\dimen@-\wd\z@
        \divide\dimen@ 3
        \@tempdima\wd\tw@ \advance\@tempdima-\scriptspace
        \divide\@tempdima 10 \advance\dimen@-\@tempdima
        \ifdim\dimen@>\z@ \dimen@0pt\fi
        \rel@kern{0.6}\kern-\dimen@
        \if#31
        \overline{\rel@kern{-0.6}\kern\dimen@\macc@nucleus\rel@kern{0.4}\kern\dimen@}%
        \advance\dimen@0.4\dimexpr\macc@kerna
        \let\final@kern#2%
        \ifdim\dimen@<\z@ \let\final@kern1\fi
        \if\final@kern1 \kern-\dimen@\fi
        \else \overline{\rel@kern{-0.6}\kern\dimen@#1}%
        \fi
    }%
    \macc@depth\@ne
    \let\math@bgroup\@empty \let\math@egroup\macc@set@skewchar
    \mathsurround\z@ \frozen@everymath{\mathgroup\macc@group\relax}%
    \macc@set@skewchar\relax
    \let\mathaccentV\macc@nested@a
    \if#31 \macc@nested@a\relax111{#1}%
    \else
    \def\gobble@till@marker##1\endmarker{}%
    \futurelet\first@char\gobble@till@marker#1\endmarker
    \ifcat\noexpand\first@char A\else \def\first@char{}%
    \fi \macc@nested@a\relax111{\first@char}%
    \fi \endgroup
}
\makeatother

\makeatletter
\newcommand*{\redefinesymbolwitharg}[1]{%
  \expandafter\let\csname ltx#1\expandafter\endcsname\csname #1\endcsname
  \@namedef{#1}{\@ifnextchar{^}{\@nameuse{#1@}}{\@nameuse{#1@}^{}}}%
  \expandafter\def\csname #1@\endcsname^##1##2{%
      \csname ltx#1\endcsname\ifx!##1!\else^{##1}\fi\mathopen{}\mathclose\bgroup\left(##2\aftergroup\egroup\right)
      }%
}
\makeatother
\redefinesymbolwitharg{sin}
\redefinesymbolwitharg{cos}

\definecolor{LightCyan}{rgb}{0.8, 0.9, 1}
\definecolor{LightGray}{rgb}{0.83, 0.83, 0.83}
\newcolumntype{b}{>{\columncolor{LightCyan}\hspace{0pt}}c}
\newcolumntype{g}{>{\columncolor{LightGray}\hspace{0pt}}c}

\usepackage{caption}
\usepackage{enumitem}
\usepackage{booktabs}
\usepackage{siunitx}
\usepackage{float}
\usepackage{algorithm}
\usepackage{algorithmicx}
\usepackage{algpseudocode}   
\newcommand{\limrefs}[2]{%
  \textcolor{firebrick}{%
    (\hyperref[#1]{L1},\,\hyperref[#2]{L2})%
  }%
}
\def\paperID{} %
\def\confName{CVPR}
\def\confYear{2026}

\title{Towards Sparse Video Understanding and Reasoning}

\author{
Chenwei Xu$^{1}$ \quad Zhen Ye$^{2}$ \quad Shang Wu$^{1}$ \quad Weijian Li$^{1}$ \quad Zihan Wang$^{1}$ \quad Zhuofan Xia$^{1}$\\
Lie Lu$^{3}$ \quad Pranav Maneriker$^{3}$ \quad Fan Du$^{3}$ \quad Manling Li$^{1}$ \quad Han Liu$^{1}$\\
{\small $^{1}$ Northwestern University \quad $^{2}$ Johns Hopkins University \quad $^{3}$ Dolby Laboratories}
}

\AtBeginDocument{%
  \crefname{theorem}{Theorem}{Theorems} 
  \crefname{proposition}{Proposition}{Propositions}
  \crefname{lemma}{Lemma}{Lemmas} 
  \crefname{corollary}{Corollary}{Corollaries}
  \crefname{definition}{Definition}{Definitions} 
  \crefname{assumption}{Assumption}{Assumptions}
  \crefname{remark}{Remark}{Remarks} 
  \crefname{problem}{Problem}{Problems}
  \crefname{property}{Property}{property}
}
\begin{document}
\maketitle
\begin{abstract}
We present \revise (\underline{Re}asoning with \underline{Vi}deo \underline{S}parsity), a multi-round agent for video question answering (VQA).
Instead of uniformly sampling frames, \revise selects a small set of informative frames, maintains a summary-as-state across rounds, and stops early when confident.
It supports proprietary vision-language models (VLMs) in a ``plug-and-play'' setting and enables reinforcement fine-tuning for open-source models.
For fine-tuning, we introduce EAGER (Evidence-Adjusted Gain for Efficient Reasoning), an annotation-free reward with three terms:
(1) Confidence gain: after new frames are added, we reward the increase in the log-odds margin between the correct option and the strongest alternative;
(2) Summary sufficiency: at answer time we re-ask using only the last committed summary and reward success;
(3) Correct-and-early stop: answering correctly within a small turn budget is rewarded.
Across multiple VQA benchmarks, \revise improves accuracy while reducing frames, rounds, and prompt tokens, demonstrating practical sparse video reasoning.
Project page: \url{https://sparsevideounderstanding.github.io}.

\end{abstract}

\section{Introduction}
Video understanding is challenging because video data are high-dimensional, temporally redundant, and semantically intricate.
Recent progress in large language models (LLMs) has accelerated video understanding research.
Many recent works~\cite{alayrac2022flamingo, bai2025qwen2_5_vl, li2024llavaonevision, wang2024videoagent, wang2025videotree, song2024moviechat} have made steady progress toward steering vision-language models (VLMs) to address these challenges.
These approaches exploit LLMs' long-context reasoning capabilities to improve question answering over video~\cite{ye2025re, wang2025lvbench, mangalam2023egoschema}.
In practice, videos are typically represented as a sequence of uniformly sampled frames.  
There are two major paradigms for integrating LLMs with those frames.  
The first involves using video captioning models to convert frames into textual descriptions, then utilizing LLMs~\cite{kahatapitiya2025language, wang2024videoagent, wang2025videotree} to perform analysis in textual space. 
However, this method may overlook fine-grained visual details inherently present in individual frames.  
To mitigate this limitation, the second directly integrates visual inputs into LLMs via pretrained vision encoders~\cite{zhai2023sigmoid}, forming vision-language models~\cite{li2024llavaonevision, liu2024llavanext, bai2025qwen2_5_vl, beyer2024paligemma}. 
However, these methods select frames uniformly, which still has two limitations, as shown in \citet{wang2025videotree}:
\phantomsection\label{lim:L1}
\textbf{\textcolor{firebrick}{(L1)} Information Overload}:
Long videos inherently exhibit substantial temporal redundancy. 
Too many redundant frames can overwhelm LLMs and hinder both reasoning and efficiency. 
\textbf{\textcolor{firebrick}{(L2)} Insufficient Key Information Awareness}:
\phantomsection\label{lim:L2}
Video content is hierarchically and temporally structured; without identifying semantically salient frames across scales, LLMs often miss critical cues for accurate reasoning.
Both limitations stem from semantic sparsity. Only a small number of frames are relevant to a given question.

\begin{figure}
    \centering
    \includegraphics[width=0.8\linewidth]{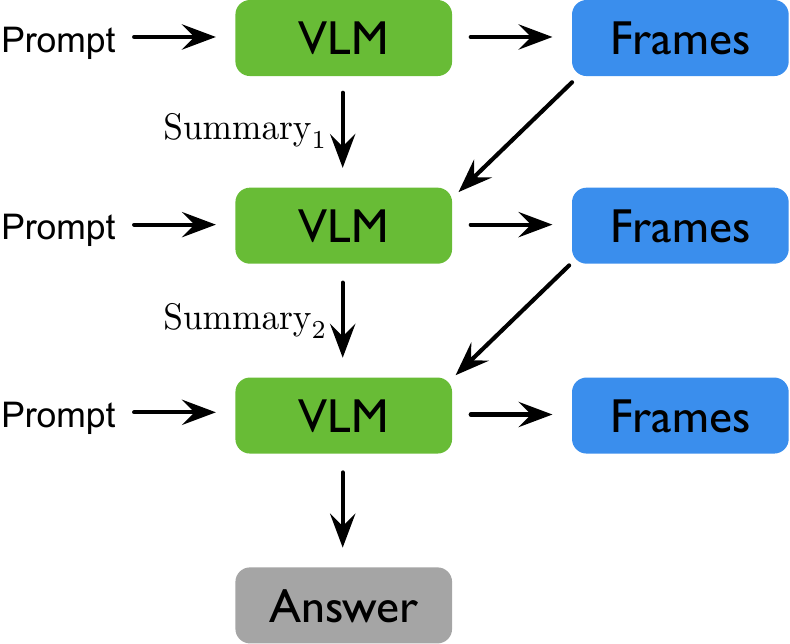}
    \vspace{-0.1in}
    \caption{\textbf{Summary-as-State.} \revise operates analogously to a recurrent neural network: it maintains a state that propagates information from previous turns to the VLM.}
    \label{fig:rnn}
    \vspace{-0.2in}
\end{figure}

To address these challenges, we introduce \revise (\underline{Re}asoning with \underline{Vi}deo \underline{S}parsity), a multi-round agent for video question answering (VQA).
Rather than processing a fixed set of uniformly sampled frames, \revise iteratively 
(i) selects a small batch of frames most likely to reduce uncertainty about the answer, 
(ii) updates a concise summary of previous-round conversations (summary-as-state) to reduce information overload, 
and (iii) stops early once the accumulated evidence is sufficient to answer. 
Conceptually, our summary-as-state is inspired by the hidden state in recurrent neural networks (e.g., LSTMs)~\cite{hochreiter1997long}, as in \Cref{fig:rnn}: it is a compact, continually updated memory that carries forward only task-critical information, informs what to ``attend to next'', and regularizes reasoning to remain faithful to accumulated evidence. 
This recurrent, stateful formulation reduces information overload \limautoref{L1} by concentrating the visual context into non-redundant text-based evidence.
Hence, this stateful design counters semantic sparsity by progressively accumulating a compact set of query-supporting frames in the persistent summary.

\begin{figure*}
    \centering
    \includegraphics[width=\linewidth]{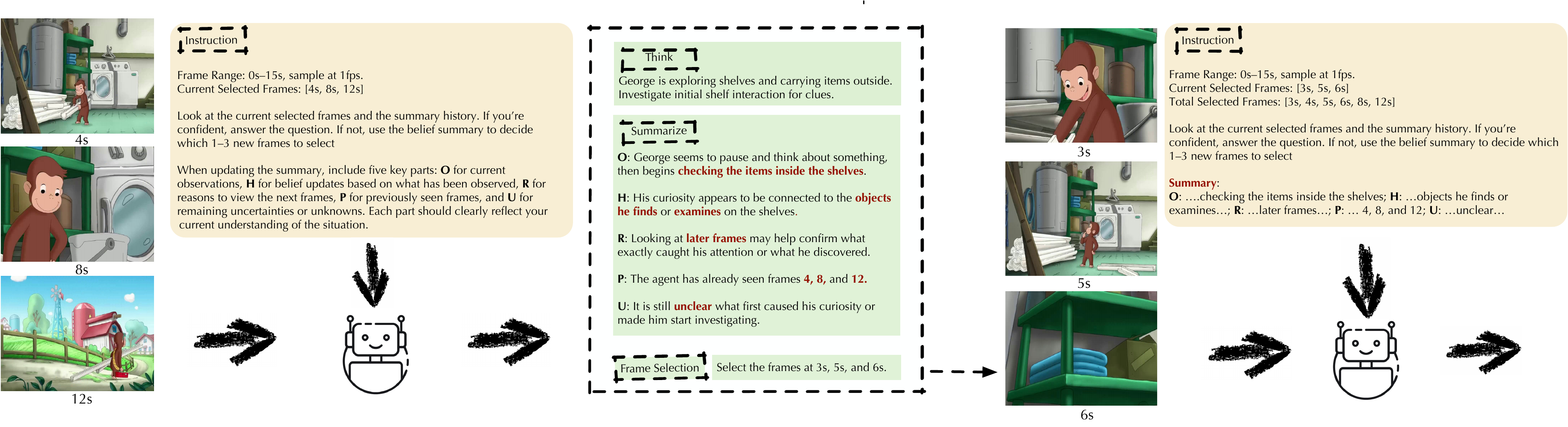}
    \vspace{-0.1in}
\caption{ \footnotesize
 \textbf{Overview of \revise: multi-round reasoning and adaptive frame selection.}  
Given an initial set of frames and a question, the VLM agent infers the video context to update the summary and selects relevant frames based on its reasoning.
In the next round, the agent reasons over the selected frames and the updated summary to generate the final answer.}
    \label{fig:high-level}
    \vspace{-0.2in}
\end{figure*}

To improve key information awareness in VLMs \limautoref{L2}, \revise uses a structured summary-as-state that explicitly tracks what has been observed, how beliefs are updated, what remains uncertain, and why additional evidence is needed.
This persistent state is compact, updated each round, and is the only information carried across turns, enabling the agent to recall prior evidence and request frames that target specific informational gaps.
We show a detailed example in \Cref{fig:high-level}.
Instead of reprocessing long conversation histories or large sets of previously seen frames, the agent continually updates this compact summary with only the verified, task-relevant evidence accumulated so far.
By grounding each decision on this evolving state rather than raw past outputs, \revise achieves coherent, uncertainty-aware multi-round reasoning while avoiding redundant processing, reducing token usage, and keeping the reasoning focused on the information that matters.
\revise operates in a fully ``plug-and-play'' manner: it wraps around any existing VLM without modifying its parameters or inference pipeline.
Moreover, for open-source VLMs, \revise can be paired with verifier-guided reinforcement fine-tuning~\cite{wang2025ragen, guo2025deepseek}, further strengthening the model's ability to identify question-critical content, reduce frame redundancy, and execute more accurate and efficient video reasoning.

Our contribution is twofold:
(I) Methodologically, we propose \revise, a novel framework for question-aware video understanding.
\revise addresses the limitations in \limautoref{L1} and \limautoref{L2} by interactively selecting informative frames through a multi-round reasoning process.
Its behavior is governed by two key components:
(1) a multi-round conversation module that progressively gathers evidence and writes an evolving ``summary-as-state'' capturing the verified information across turns; and
(2) a structured summary-as-state that records observations, belief updates, uncertainties, and selection rationale, and is the only information carried across rounds for stable, uncertainty-aware reasoning.
\revise is lightweight and modular: it wraps around any existing VLM in a plug-and-play manner without parameter updates.
Further, when paired with verifier-guided reinforcement fine-tuning, \revise strengthens open-source VLMs' ability to locate question-critical frames and reason over long videos more efficiently.
(II) Experimentally, we conduct extensive evaluations across diverse video understanding benchmarks, spanning short clips to hour-long videos.
In the plug-and-play setting, \revise improves efficiency while matching or exceeding strong proprietary VLM baselines.
With reinforcement fine-tuning, \revise further boosts the performance of open-source VLMs, achieving higher accuracy with significantly fewer frames, fewer rounds, and fewer prompt tokens.
Together, these results demonstrate that \revise delivers practical and scalable sparse video reasoning.

\section{Related Works}
\textbf{VLMs for Video Understanding.}
Several recent approaches extend image-centric vision-language models~\cite{liu2023visual, bai2023qwenvl, wang2024cogvlm, ye2025re} to video by treating videos as sequences of frames.
Many methods~\cite{munasinghe2023pg, lin2024video, liu2024llavanext, korbar2024text, weng2024longvlm, maaz2024video, zhang2023videollama, chen2023video, li2025videochat, jin2024chat, he2024ma, li2024llms, wang2024lstp, li2024mvbench, yu2024crema} train vision-language models by connecting a vision encoder to a large language model through a lightweight adapter. 
Training-free methods~\cite{kahatapitiya2025language, fan2024videoagent, wang2024videoagent, suris2023vipergpt, wang2024vamos, ko2023large, wang2023lifelongmemory, wang2025videotree} integrate captioners with LLMs to support video understanding without full retraining.
These pipelines also support interactive selection and open-ended relational reasoning for video QA~\cite{wang2024videoagent,luo2023open}.
Specifically, LLoVi~\cite{zhang2024simple} generates short-term video captions using a visual encoder and prompts an LLM to summarize and answer user queries. 
VideoAgent~\cite{wang2024videoagent} introduces an LLM-driven multi-round frame selection strategy based on captioned content.
Recent VLMs such as LLaVA~\cite{liu2024llavanext, li2024llavaonevision, liu2023visual} and QwenVL~\cite{bai2023qwenvl, wang2024qwen2vl, bai2025qwen2_5_vl} integrate stronger vision-language reasoning capabilities and reduce reliance on external captioners.
\revise combines agent-based search from VideoAgent with the direct visual reasoning abilities of modern VLMs.

\noindent \textbf{Adaptive Frame Selection for Video LLMs.}
A growing body of work addresses the challenge of selecting informative frames from videos to improve VLM efficiency and accuracy.
\emph{Training-free approaches} avoid any parameter updates: MDP3~\cite{sun2025mdp3} formulates list-wise frame selection as a Markov decision process solved without training; Q-Frame~\cite{zhang2025q} uses CLIP-based text-image matching with Gumbel-Max sampling for query-aware selection; FRAG~\cite{huang2025frag} asks the model itself to score each frame's relevance; and FOCUS~\cite{zhu2025focus} casts keyframe selection as a combinatorial pure-exploration bandit problem.
\emph{Learned selectors} train lightweight policies: Frame-Voyager~\cite{yu2025framevoyager} learns to query task-relevant frames for video LLMs; Flexible Frame Selection~\cite{buch2025flexible} proposes a differentiable top-$k$ selection operation trained end-to-end; Adaptive Keyframe Sampling~\cite{tang2025adaptive} learns to sample keyframes tailored to long videos; M-LLM~\cite{hu2025m} leverages multimodal LLMs to guide frame selection; and K-frames~\cite{yao2025k} performs scene-driven any-$k$ keyframe selection.
\emph{RL-based methods} optimize frame selection policies via reinforcement learning: TSPO~\cite{tang2026tspo} learns a temporal sampling policy with policy optimization; ReFoCUS~\cite{lee2025refocus} uses reinforcement-guided frame optimization for contextual understanding; FrameMind~\cite{ge2025framemind} enables frame-interleaved reasoning via GRPO; and FrameThinker~\cite{he2025framethinker} combines SFT with GRPO for multi-turn frame spotlighting.
\emph{Iterative agent-based approaches} perform multi-round frame selection guided by reasoning: Active Video Perception~\cite{wang2025active} employs a plan-observe-reflect loop; A.I.R.~\cite{zou2025air} uses adaptive, iterative, reasoning-based selection; and VideoBrain~\cite{zou2026videobrain} deploys dual complementary agents with behavior-aware rewards.
Unlike prior work that typically performs a single-pass selection, \revise maintains a persistent \emph{summary-as-state} across rounds and couples it with EAGER reward for RL fine-tuning, enabling question-aware, iterative frame selection that remains compatible with any VLM in a plug-and-play manner.

\noindent \textbf{VLM Reasoning and Multi-round Conversation.}
Recent progress in vision-language models (VLMs) has shifted from single-turn processing to interactive, reasoning-centric systems capable of sustaining multi-round dialog. Early efforts~\cite{yang2023set,yang2023dawn,zeng2022socratic} enhanced interaction through prompt engineering and external APIs, bypassing the need for fully end-to-end architectures.
More recent approaches incorporate explicit reasoning, enabling VLMs to infer answers based on implied visual information. For instance, DetGPT~\cite{pi2023detgpt} performs object detection through high-level instructions rather than predefined class labels. GPT4RoI~\cite{zhang2023gpt4roi} uses spatial boxes to focus attention on specific regions, improving alignment between vision and language. Similarly, LISA~\cite{lai2024lisa} augments the mask decoder in SAM~\cite{kirillov2023segment} with a learned embedding prompt, enabling high-level visual reasoning when paired with LLaVA~\cite{liu2023visual}.
Complementing these architectural advances, reinforcement learning techniques have emerged as effective tools for enhancing multi-step reasoning. PPO-based ReFT~\cite{luong2024reft} rewards correct chains of thought, while DeepSeek-R1~\cite{guo2025deepseek} introduces step-wise rewards for logical soundness. DeepSeek-R1-Zero~\cite{guo2025deepseek} demonstrates that outcome-only rewards can suffice when reasoning is self-verifiable. RAGEN~\cite{wang2025ragen} further shows that intermediate rewards are essential in preventing dialog agents from adopting shallow heuristics. Finally, self-revision models like Draft-Edit~\cite{kumar2024training}, S2R~\cite{ma2025s2r}, and token-level reward ``dancing''~\cite{li2025dancing} underscore the value of iterative feedback for improving reasoning depth.
Overall, these studies show that feedback and revision can drive multi-turn reasoning, ranging from reward shaping to simplified feedback signals~\cite{wang2025ragen,liu2025simple,kumar2024training,ma2025s2r,li2025dancing}.
Together, these developments point toward a new generation of VLMs that not only ground visual input accurately but also engage in self-correcting, multi-round reasoning to produce coherent and reliable answers.

\section{Methodology}
\begin{figure*}
    \centering
    \includegraphics[width=0.85\textwidth]{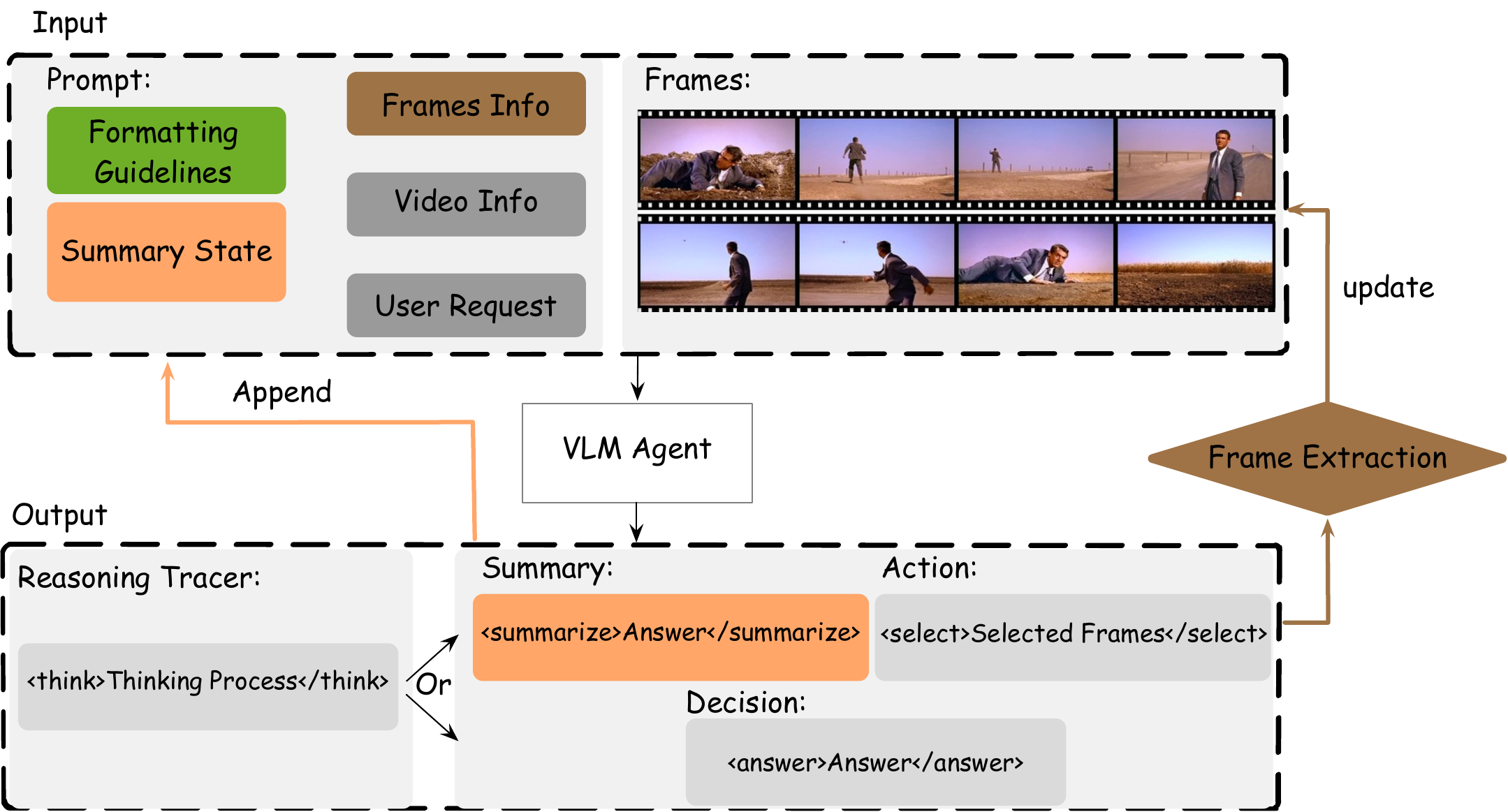}
    \vspace{-0.05in}
    \caption{\footnotesize \textbf{\revise}. 
\revise consists of three components: multi-round conversation, a structured output protocol, and a summary-as-state.
Each round, the VLM agent receives (i) the entire conversation history, (ii) the current prompt, and (iii) the chosen video frames, annotated with their timestamps and the video's total frame count.
In the first round, a formatting guideline is also provided.
Each round, the VLM first emits a \texttt{<think>} reasoning trace, then either commits a \texttt{<summarize>} state ($P/O/H/U/R$) together with a \texttt{<select>} request (non-final round) or returns an \texttt{<answer>} (final round); the committed summary carries the persistent state across rounds.
\revise then extracts the new frames and starts the next round with updated prompt and history.
The conversation ends when the VLM produces a valid answer or the maximum number of rounds is reached.
    }\vspace{-0.1in}
    \label{fig:revise}
\end{figure*}

As illustrated in \Cref{fig:revise}, \revise couples multi-round interaction with an explicit, structured summary-as-state.
It targets two limitations of current VLMs in video understanding~\cite{wang2025videotree}: \limautoref{L1} information overload and \limautoref{L2} insufficient key-information awareness.
We cast video understanding as an iterative, question-aware frame-selection problem that admits only frames most likely to support the query.
\revise mitigates \limautoref{L1} via budgeted multi-round frame selection and improves \limautoref{L2} by tracking observations, belief updates, uncertainties, and selection rationale in a persistent summary.

\noindent \textbf{Problem Formulation.}
Given a video $V=\{x_i\}_{i=0}^{L-1}$ consisting of $L$ frames, with the frame at time $i$ denoted by $x_i$, and a user prompt $p$, the goal is to produce an answer $a$ with a VLM agent $\pi_\theta$ while respecting a maximum context budget $K$.
Let $c(x_i)$ denote the model-specific visual token cost of frame $x_i$, and $C(F)=\sum_{x\in F} c(x)$ the cost of a subset $F\subseteq V$.
Instead of processing all frames (which may violate $K$), we construct the visual context \emph{iteratively} and maintain a compact, persistent \emph{``summary-as-state''}.

We model the interaction over at most $T$ rounds.
Let $S_t=\bigcup_{j=1}^{t} F_j$ be the set of all frames admitted up to round $t$ (with $S_0=\varnothing$).
Let $p_t$ denote the prompt at round $t$, constructed from the original prompt $p$, the previous summary $z_{t-1}$, and the shown frames $F_t$ (with timestamps and basic video meta).
The agent maintains a summary state:
\begin{equation}
z_t \;=\; (P_t,\,O_t,\,H_t,\,U_t,\,R_t),
\end{equation}
where $P_t$ (previously seen) summarizes what has already been inspected, $O_t$ (observations) records the currently observed evidence, $H_t$ (belief updates) captures how those observations update the hypothesis (without outputting the final answer letter), $U_t$ (uncertainties) enumerates remaining unknowns, and $R_t$ (reasons) states what evidence to look for next (or that the question is answered). The order is fixed as $P \rightarrow O \rightarrow H \rightarrow U \rightarrow R$.
Operationally, $z_t$ is written explicitly in the \texttt{<summarize>} field and is the \emph{only} state carried across rounds.
\textit{Each turn conditions on all previous states}: the agent has access to the set of committed summaries $\{z_j\}_{j=0}^{t-1}$; since $z_{t-1}$ is defined to be \emph{cumulative}, conditioning on $\{z_j\}_{j<t}$ is equivalent to conditioning on the latest $z_{t-1}$ alone.

At round $t$, given $(p_t, z_{t-1}, F_t)$, the agent takes a single action
\begin{equation}
a_t \in \bigl\{\textsc{Select}(Q_t),\ \textsc{Answer}(y_t)\bigr\},
\end{equation}
where $Q_t \subseteq \{0,\ldots,L-1\}\setminus S_{t-1}$ is a small set of requested frame indices (0-based).
If $a_t=\textsc{Select}(Q_t)$, the environment retrieves $F_{t+1}=\{x_i:\,i\in Q_t\}$, updates $S_{t+1}=S_{t}\cup F_{t+1}$, and the agent commits the next summary
\begin{equation}
z_t \;=\; f_\theta\!\bigl(z_{t-1},\,p_t,\,F_{t+1}\bigr) \;=\; \bigl(P_t,\,O_t,\,H_t,\,U_t,\,R_t\bigr),
\end{equation}
with $R_t$ explicitly guiding the proposal of the next $Q_{t+1}$.
If $a_t=\textsc{Answer}(y_t)$, the process stops with $a=y_t$ at stopping time $\tau\le T$.
At all times we enforce the token budget $C(S_t)+|p_t|\le K$.

Our objective is to choose the selection sequence $\{Q_t\}_{t=1}^{\tau-1}$ and stopping time $\tau$ (implicitly via the agent policy) to maximize task performance under budget:
\begin{equation}
\max_{\{Q_t\},\,\tau\le T}\ \ \mathcal{R}\!\Bigl(\pi_\theta\bigl(p,\,z_{\tau-1},\,S_{\tau-1}\bigr)\Bigr)
\quad \text{s.t.}\ \ C(S_{\tau-1})+|p_\tau|\le K.
\end{equation}
This sequential formulation replaces single-shot maximum coverage with a summary-driven, question-aware selection process: the agent repeatedly (i) reads a few frames, (ii) updates $z_t=(P,O,H,U,R)$ so it cumulatively encodes all previous states, and (iii) decides (based on $R_t$ and $U_t$) what to view next or when to answer.
We detail how \revise instantiates the multi-round framework and how the summary-as-state enables efficient, query-focused reasoning below.

\subsection{\revise Building Components}
\label{sec:module}
The core concept of \revise is to admit only frames that are relevant to the user's request while maintaining a compact state that carries verified evidence across rounds.
Accordingly, \revise comprises three modules as in \Cref{fig:revise}:
(i) a multi-round controller,
(ii) a structured response format that externalizes the summary state, and
(iii) a persistent summary-as-state.
The multi-round controller adaptively selects frames and decides when to answer, while the summary-as-state is the sole memory that accumulates and conditions future decisions.

\noindent \textbf{Multi-Round Controller.}
The agent interacts for at most $T$ rounds.
Let $F_t$ be the frames shown at round $t$ and $S_t=\bigcup_{j=1}^{t}F_j$ the union of all admitted frames ($S_0=\varnothing$).
Round $t{=}1$ starts from a small, uniformly sampled set $F_1$ and the initial prompt $p$ (the prompt includes timestamps and basic video meta).
At each subsequent round $t\!\ge\!2$, given $(p_t, z_{t-1}, F_t)$, the agent produces a single action:
\begin{equation}
  a_t \in \{\textsc{Select}(Q_t), \textsc{Answer}(y_t)\},
\end{equation}

where $Q_t\subseteq \{0,\ldots,L-1\}\setminus S_{t-1}$ requests a few new indices (0-based).
If $a_t=\textsc{Select}(Q_t)$, the environment fetches $F_{t+1}=\{x_i:\, i\in Q_t\}$, updates $S_{t+1}=S_{t}\cup F_{t+1}$, and the agent commits the next summary $z_t$.
If $a_t=\textsc{Answer}(y_t)$, the process terminates.
We enforce the token budget $C(S_t)+|p_t|\le K$ throughout.
\emph{Each turn conditions on all previous states} via the cumulative summary: by construction, $z_{t-1}$ subsumes $\{z_0,\ldots,z_{t-2}\}$, so conditioning on $z_{t-1}$ is equivalent to conditioning on the entire state history with constant memory cost.
For notational clarity, the action is produced as:
\begin{equation}
a_t=\pi_\theta(p_t, z_{t-1}, F_t).
\end{equation}

\noindent \textbf{Structured Response Format.}
To make decisions transparent and to improve the quality of summary state, every response begins with a \texttt{<think>} reasoning trace and then takes one of two formats:
\begin{description}[style=nextline, leftmargin=0pt, labelsep=0.4em, itemsep=0pt, topsep=0pt]
  \item[\textsc{Select:}]
  \texttt{<think>} \ldots \texttt{</think>}
  \texttt{<summarize>} \ldots \texttt{</summarize>}
  \texttt{<select>} \ldots \texttt{</select>}.
  \item[\textsc{Answer:}]
  \texttt{<think>} \ldots \texttt{</think>}
  \texttt{<answer>} \ldots \texttt{</answer>}.
\end{description}

\noindent The \texttt{<think>} trace exposes the per-round reasoning but is \emph{not} persisted; only the \texttt{<summarize>} state committed on \textsc{Select} rounds is carried to the next round. The final \textsc{Answer} round emits only \texttt{<think>} and \texttt{<answer>}, reusing the last committed summary. This design keeps the prompt compact, improves interpretability, and guides query-aware selection and early stopping in subsequent rounds.

\noindent \textbf{Summary-as-State.}
The persistent state is written explicitly in \texttt{<summarize>} as
\begin{equation}
z_t=(P_t, O_t, H_t, U_t, R_t),
\end{equation}
where $P_t$ (Previously seen) summarizes what has already been inspected, $O_t$ (Observations) records what was just seen, $H_t$ (belief \underline{H}ypotheses/updates) captures how those observations change the current hypothesis, $U_t$ (Uncertainties) enumerates remaining unknowns, and $R_t$ (Reasons) justifies which frames to view next (or indicates the question is answered).
By carrying forward only $z_t$ rather than raw histories, \revise (i) avoids information overload by not re-admitting redundant context and (ii) improves key-information awareness by explicitly tracking what has been observed ($O$), how beliefs change ($H$), what remains uncertain ($U$), and what to request next and why ($R$).
The $R_t$ component directly informs the next proposal $Q_{t+1}$, and the stability of $H_t$/$U_t$ provides a natural signal for when to answer.
At each round, the policy first reasons in a free-form \texttt{<think>} trace to decide what to view next, and then distills the outcome into the structured state: the $(H_t,U_t,R_t)$ fields in \texttt{<summarize>} record the committed thinking, i.e., hypothesis updates, remaining uncertainties, and reasons for the next action.
Unlike the transient \texttt{<think>} trace, this distilled state is the only reasoning carried across rounds.

\subsection{Plug-and-Play}

\revise treats any proprietary vision-language model as a frozen black-box module and communicates with it solely through its public inference interface (e.g., an API).
All operations, including multi-round conversation, adaptive frame selection, structured summary updates, and validity enforcement, run externally within the framework, so no parameter updates are required. 
This design lets \revise plug into closed-source systems as is, automatically orchestrating iterative multi-round interactions while leaving the model's original weights and vision-processing capabilities unchanged.

\subsection{Reinforcement Fine-Tuning}
Following \citet{wang2025ragen}, we cast the \revise multi-round interaction as a finite-horizon MDP $\mathcal{M}=\langle\mathcal{S},\mathcal{A},\mathcal{T},r,\gamma\rangle$ with horizon $T$.
At round $t$, the state is
\begin{equation}
s_t=(p_t,\, z_{t-1},\, S_{t-1}),
\end{equation}
where $p_t$ is the current prompt (with formatting/meta), $z_{t-1}$ is the cumulative summary-as-state committed in the previous round, and $S_{t-1}=\bigcup_{j=1}^{t-1}F_j$ is the set of admitted frames so far.
The action space is
\begin{equation}
a_t \in \{\textsc{Select}(Q_t),\; \textsc{Answer}(y_t)\},
\end{equation}
where $Q_t\subseteq\{0,\ldots,L-1\}\setminus S_{t-1}$ requests new frame indices (0-based) and $y_t$ is an answer.
The transition updates $(S_t,z_t)$:
if $a_t=\textsc{Select}(Q_t)$, the environment returns $F_{t+1}=\{x_i:\,i\in Q_t\}$, sets $S_{t+1}=S_{t}\cup F_{t+1}$, and the agent commits the next summary
$z_t=f_\theta(z_{t-1},p_t,F_{t+1})$; if $a_t=\textsc{Answer}(y_t)$, the episode terminates at $\tau\le T$ with answer $a=y_t$.
We optimize the expected return
\begin{equation}
J(\theta)=\mathbb{E}_{H\sim\pi_\theta}\!\left[\sum_{t=1}^{\tau}\gamma^{t-1} r_t\right],
\end{equation}
and decompose $\pi_\theta$ into token-level likelihoods to remain compatible with autoregressive VLMs.

\noindent\textbf{Reward Design: EAGER.}
We design a dense, annotation-free reward that aligns the policy with efficient, summary-driven reasoning.
Let $\mathcal{Y}$ be the answer set (e.g., MCQ options) and $y^\star\in\mathcal{Y}$ the correct label.
Define a (temperature-calibrated) log-odds margin under the current context
\begin{equation}
m_t \;=\; \log p_\theta\!\big(y^\star \mid p_t, z_{t-1}, S_t\big)\;-\;\max_{y\neq y^\star}\log p_\theta\!\big(y \mid p_t, z_{t-1}, S_t\big),
\end{equation}
computed at each decision state (before taking action $a_t$).
EAGER comprises three parts:

\emph{(i) Confidence gain.}
Reward only evidence that \emph{truly} increases certainty after adding new frames:
\begin{equation}
r^{\text{conf}}_t \;=\; \big[m_{t+1} - m_t\big]_+\quad\text{(applies on \textsc{Select})}.
\end{equation}

\emph{(ii) Summary sufficiency.}
At answer time, re-ask using \emph{summary-only} to encourage faithful, compact state:
\begin{equation}
\begin{aligned}
r^{\text{sum}}_t
&=\mathds{1}\!\left[\arg\max_{y\in\mathcal{Y}} p_\theta\!\big(y \mid p_\tau, z_{\tau-1}\big) = y^\star\right]\\
&\quad\text{(applies on \textsc{Answer})}.
\end{aligned}
\end{equation}

\emph{(iii) Correct-and-early stop.}
We reward answering correctly within a small turn budget $T_{\text{stop}}$:
\begin{equation}
r^{\text{stop}}_t =
\begin{cases}
1 + \beta\,\big[T_{\text{stop}}-\tau\big]_+, & a_\tau=\textsc{Answer}(y^\star)\ \text{and}\ \tau \le T_{\text{stop}},\\
0, & \text{otherwise},
\end{cases}
\end{equation}
applied at the final step. We also include a small structural bonus $r^{\text{format}}_t=\alpha\cdot\mathds{1}[\text{valid format}]$ for emitting valid tags (a \texttt{<think>} trace followed by either \texttt{<summarize>} and \texttt{<select>}, or \texttt{<answer>}, with no extra text).

The per-step reward is
\begin{equation}
r_t \;=\; \lambda_1\, r^{\text{conf}}_t \;+\; \lambda_2\, r^{\text{sum}}_t \;+\; \lambda_3\, r^{\text{stop}}_t \;+\; r^{\text{format}}_t,
\end{equation}
and the episode return is $R(H)=\sum_{t=1}^{\tau}\gamma^{t-1} r_t$.
EAGER requires only answer labels (for $y^\star$) and model scores; it does \emph{not} use frame-level annotations.
\begin{table*}[t]
\centering
\small
\setlength{\tabcolsep}{3pt}
\renewcommand{\arraystretch}{1.05}
\caption{\footnotesize \textbf{Comparison of VLMs across fine-grained video reasoning categories.} 
We report accuracy (\%) per category of baseline results, quoted from \citet{han2025videoespresso} and \revise under the same protocol of baselines; 
\#Frames reports either the average frames processed per video (ours) or a uniform sampling rate when shown as \texttt{FPS{=}$k$}; \emph{Param} is the backbone size. 
Underlines mark the better score within each backbone pair; bold indicates the group best per column.
}
\vspace{-0.1in}
\resizebox{\textwidth}{!}{
\begin{tabular}{lcccccccccccccccccc}
\toprule
\textbf{Model} & \textbf{\#Frames} & \textbf{Param} &
\textbf{Narra.} & \textbf{Event} & \textbf{Ingre.} & \textbf{Causal} &
\textbf{Theme} & \textbf{Conte.} & \textbf{Influ.} & \textbf{Role} &
\textbf{Inter.} & \textbf{Behav.} & \textbf{Emoti.} &
\textbf{Cook.} & \textbf{Traff.} & \textbf{Situa.} & \textbf{Avg.} \\

\midrule
\multicolumn{18}{l}{\textbf{Open-source VLMs}} \\

LLaVA-1.5~\cite{liu2024improved} & 4 & 7B &
32.3 & 21.3 & 19.4 & 17.1 & 26.2 & 20.2 & 36.1 & 33.3 &
21.0 & 21.1 & 20.0 & 35.8 & 16.7 & 18.0 & 24.2 \\

LLaVA-N-Inter~\cite{li2024llava} & FPS=1 & 7B &
24.2 & 23.6 & 26.5 & 19.2 & 31.1 & 32.1 & 31.9 & 17.5 &
24.2 & 21.1 & 26.2 & 30.2 & 13.3 & 20.0 & 24.4 \\

LongVA-DPO~\cite{zhang2024long} & 128 & 7B &
35.5 & 14.9 & 16.3 & 19.0 & 34.4 & 22.0 & 37.5 & 23.8 &
29.0 & 22.8 & 20.0 & 37.7 & 16.7 & 12.0 & 24.4 \\

	mPLUG-Owl3~\cite{ye2024mplug} & FPS=1 & 7B &
	30.6 & 23.6 & 20.4 & 22.3 & 37.7 & 29.4 & 48.6 & 34.9 &
	\textbf{30.6} & 24.6 & 27.7 & 24.5 & 13.3 & 24.0 & 28.0 \\

LLaVA-N-Video~\cite{li2024llava} & FPS=1 & 7B &
31.2 & 20.2 & 16.2 & 17.6 & 36.5 & 32.7 & 30.6 & 24.5 &
26.4 & 24.5 & 34.7 & 20.8 & 20.3 & 17.0 & 25.2 \\

	VideoEspresso~\cite{han2025videoespresso} & 2.36 & 8.5B &
	\textbf{45.2} & 27.0 & 33.7 & 26.1 &
	39.3 & 36.7 & \textbf{55.6} & 41.3 &
	\textbf{30.6} & \textbf{29.8} & 30.8 & 35.8 &
	20.0 & 26.0 & 34.1 \\

\cellcolor{LightGray} InternVL2~\cite{chen2024internvl} & FPS=1 & 8B &
\cellcolor{LightGray} 33.9 & \cellcolor{LightGray} 24.1 & \cellcolor{LightGray} 27.6 & \cellcolor{LightGray} \underline{24.4}&\cellcolor{LightGray}  \underline{42.6} & \cellcolor{LightGray} 33.0 & \cellcolor{LightGray} 45.8 & \cellcolor{LightGray} 28.6 &
\cellcolor{LightGray} \underline{19.4} & \cellcolor{LightGray} 22.8 & \cellcolor{LightGray} 21.5 & \cellcolor{LightGray} 34.0 & \cellcolor{LightGray} 20.0 & \cellcolor{LightGray} \underline{24.0} & \cellcolor{LightGray} 28.7 \\

	\cellcolor{LightCyan} \underline{InternVL2 + ReViSe} & 2.87 & 8B &
	\cellcolor{LightCyan}\underline{35.5} &
	\cellcolor{LightCyan}{\underline{29.6}} &
	\cellcolor{LightCyan}{\underline{36.0}} &
	\cellcolor{LightCyan}{21.3} &
	\cellcolor{LightCyan}{39.3} &
	\cellcolor{LightCyan}{33.0} &
	\cellcolor{LightCyan}{36.1} &
	\cellcolor{LightCyan}{\underline{31.7}} &
	\cellcolor{LightCyan}{18.3} &
	\cellcolor{LightCyan}\underline{24.6} &
	\cellcolor{LightCyan}\underline{36.9} &
	\cellcolor{LightCyan}{\underline{\textbf{40.9}}} &
	\cellcolor{LightCyan}{\underline{\textbf{42.9}}} &
	\cellcolor{LightCyan}{23.4} &
	\cellcolor{LightCyan}{\underline{32.1}} \\

\cellcolor{LightGray} Qwen2-VL~\cite{wang2024qwen2vl} & FPS=1 & 7B &
\cellcolor{LightGray} 27.4 & \cellcolor{LightGray} 23.0 &\cellcolor{LightGray}  24.5 & \cellcolor{LightGray} 23.5 & \cellcolor{LightGray} 29.5 & \cellcolor{LightGray} 31.2 & \cellcolor{LightGray} \underline{47.2} & \cellcolor{LightGray} 31.7 &
\cellcolor{LightGray} 22.6 & \cellcolor{LightGray} 28.1 & \cellcolor{LightGray} 40.0 & \cellcolor{LightGray} 22.6 &\cellcolor{LightGray}  30.0 & \cellcolor{LightGray} 18.0 & \cellcolor{LightGray} 28.5 \\

	\cellcolor{LightCyan} \underline{Qwen2-VL + ReViSe} & 6.25 & 7B &
	\cellcolor{LightCyan}{\underline{39.2}} &
	\cellcolor{LightCyan}{\underline{\textbf{39.5}}} &
	\cellcolor{LightCyan}{\underline{\textbf{38.1}}} &
	\cellcolor{LightCyan}{\underline{\textbf{33.3}}} &
	\cellcolor{LightCyan}{\underline{\textbf{50.8}}} &
	\cellcolor{LightCyan}{\underline{\textbf{46.7}}} &
	\cellcolor{LightCyan}{40.3} &
	\cellcolor{LightCyan}\underline{\textbf{50.0}} &
	\cellcolor{LightCyan}\underline{25.8} &
	\cellcolor{LightCyan}{28.1} &
	\cellcolor{LightCyan}{\underline{\textbf{43.1}}} &
	\cellcolor{LightCyan}\underline{33.3} &
	\cellcolor{LightCyan}\underline{36.7} &
	\cellcolor{LightCyan}\underline{\textbf{38.8}} &
	\cellcolor{LightCyan}{\underline{\textbf{37.8}}} \\

\midrule
\multicolumn{18}{l}{\textbf{Closed-source VLMs}} \\

Qwen-VL-Max~\cite{bai2023qwenvl} & FPS=3 & -- &
33.9 & 22.4 & 23.5 & 21.4 & 26.2 & 30.3 & 41.7 & 30.2 &
27.4 & 26.3 & 20.0 & 20.8 & 16.7 & 24.0 & 26.0 \\

\cellcolor{LightGray} GPT-4o~\cite{gpt4o} & FPS=3 & -- &
\cellcolor{LightGray} 32.3 & \cellcolor{LightGray} 16.7 & \cellcolor{LightGray} 25.5 & \cellcolor{LightGray} 22.8 &\cellcolor{LightGray}  32.8 & \cellcolor{LightGray} 27.5 &\cellcolor{LightGray}  37.5 & \cellcolor{LightGray} 28.6 &
\cellcolor{LightGray} 24.2 & \cellcolor{LightGray} 19.3 &\cellcolor{LightGray}  30.8 &\cellcolor{LightGray}  30.2 &\cellcolor{LightGray}  20.0 & \cellcolor{LightGray} 22.0 &\cellcolor{LightGray}  26.4 \\

	\cellcolor{LightCyan} GPT-4o + ReViSe & 7.99 & -- &
	\cellcolor{LightCyan}\underline{\textbf{51.9}} &
	\cellcolor{LightCyan}\underline{\textbf{46.5}} &
	\cellcolor{LightCyan}\underline{\textbf{55.1}} &
	\cellcolor{LightCyan}\underline{\textbf{44.0}} &
	\cellcolor{LightCyan}\underline{\textbf{54.1}} &
	\cellcolor{LightCyan}\underline{\textbf{53.2}} &
	\cellcolor{LightCyan}\underline{\textbf{48.6}} &
	\cellcolor{LightCyan}{\underline{\textbf{50.8}}} &
	\cellcolor{LightCyan}{\underline{\textbf{40.3}}} &
	\cellcolor{LightCyan}{\underline{\textbf{49.1}}} &
	\cellcolor{LightCyan}{\underline{\textbf{50.8}}} &
\cellcolor{LightCyan}{\underline{\textbf{58.5}}} &
\cellcolor{LightCyan}{\underline{\textbf{53.3}}} &
\cellcolor{LightCyan}{\underline{\textbf{58.0}}} &
\cellcolor{LightCyan}{\underline{\textbf{48.9}}} \\
\bottomrule
\end{tabular}
}
\label{tab:videoespresso}
\vspace{-0.1in}
\end{table*}

\noindent\textbf{Policy Optimization.}
We optimize $\pi_\theta$ with GRPO~\cite{shao2024deepseekmath}.
Each iteration, starting from $F_1$ and $p_1$, we sample $G$ trajectories $\{H^i\}_{i=1}^{G}$, compute scalar returns $R(H^i)=\sum_{t}\gamma^{t-1} r_t^i$, and standardize them to obtain a trajectory-level advantage
\begin{equation}
\hat A_i=\frac{R(H^i)-\mathrm{mean}(R(H^\cdot))}{\mathrm{std}(R(H^\cdot))},
\end{equation}
which is shared across all tokens of trajectory $i$.
Let $H^{i,(n)}$ be the $n$-th token and $N_i$ the token count in $H^i$.
The GRPO objective is
\begin{align}
\begin{aligned}
J_{\text{GRPO}}(\theta)
&= \frac{1}{G}\sum_{i=1}^{G}\frac{1}{N_i}\sum_{n=1}^{N_i}
\min\!\Big(
    \rho_{i,n}\,\hat A_i,\;
\\[-2pt]
&\qquad\qquad\qquad
    \mathrm{clip}(\rho_{i,n},1-\epsilon,1+\epsilon)\,\hat A_i
\Big)
\end{aligned}
\end{align}
\vspace{-0.2in}
\begin{align}
\rho_{i,n}
&= \frac{\pi_\theta(H^{i,(n)}\mid H^{i,<n})}{\pi_{\text{old}}(H^{i,(n)}\mid H^{i,<n})}.
\end{align}

\section{Experimental Studies}

We integrate \revise with both proprietary and open-source VLMs and test it against several baselines.
Specifically, we evaluate Qwen2-VL-7B~\cite{wang2024qwen2vl}, Qwen2.5-VL (3B/7B)~\cite{bai2025qwen2_5_vl}, InternVL2-8B~\cite{chen2024internvl}, and GPT-4o~\cite{gpt4o}.

\noindent \textbf{Datasets.} 
We report results on three complementary video-QA benchmarks that probe different temporal scales and reasoning demands.
\textit{VideoEspresso}~\cite{han2025videoespresso} is a large-scale, chain-of-thought (CoT) video reasoning corpus built with a core-frame selection pipeline and multimodal CoT evidence. The benchmark organizes evaluation into 14 tasks spanning causal, temporal, spatial, and high-level narrative reasoning, and emphasizes answering from sparse core frames rather than full streams.
\textit{NExT-QA}~\cite{xiao2021next} targets causal and temporal action reasoning with both multiple-choice and open-ended QA. Videos average 44 seconds, and questions are stratified into causal (48\%), temporal (29\%), and descriptive (23\%) types. We report accuracy for multiple-choice following the official split.
\textit{EgoSchema}~\cite{mangalam2023egoschema} is a long-form egocentric benchmark with $>$5{,}000 five-choice multiple-choice questions over 3-minute clips. 

\noindent \textbf{Settings.}
We follow a fixed interaction budget for all multi-turn methods. 
We set \texttt{max\_frames\_per\_round} to 3, i.e., each evidence-gathering step can inspect at most three frames 
($\max(F_1,\dots,F_T)=3$), and limit the \texttt{max\_rounds}=4 ($T=4$). 
For ablation study, we use VideoEspresso~\cite{han2025videoespresso} with Qwen2.5-VL-7B~\cite{bai2025qwen2_5_vl}, for efficiency.  
Evaluation reports both answer accuracy and the total frame budget consumed by each method. 
Unless otherwise specified, all VLMs are queried with temperature~0.2, a maximum response length of 256 tokens, and top-$p$ sampling with $p=0.9$.
All VLM experiments are conducted with 4$\times$80G A100.

\subsection{Plug-and-Play}
\label{sec:exp_pnp}
\revise improves any model's video understanding abilities in a ``plug-and-play'' fashion, without any need to fine-tune the model weights.
We compare \revise with both the proprietary and open-source models across various multiple-choice VQA datasets with several baselines. 
Our results show that \revise can achieve comparable performance with far fewer frames.

\noindent \textbf{Improvements with Baselines on VideoEspresso Dataset.} As in \Cref{tab:videoespresso}, \revise consistently improves the backbones across both open- and closed-source settings while using only a single-digit number of frames on average. 
With open-source models, Qwen2-VL~\cite{wang2024qwen2vl} + \revise raises the average from 28.5 to 37.8 (+9.3) and InternVL2~\cite{chen2024internvl} + \revise from 28.7 to 32.1 (+3.4). On the closed-source side, GPT-4o~\cite{gpt4o} + \revise improves the average from 26.4 to 48.9 (+22.5) and achieves the best score in 13 of 14 fine-grained categories, outperforming prior systems such as VideoEspresso.
Remarkably, these improvements require only 2.87, 6.25, and 7.99 frames per video for InternVL2, Qwen2-VL, and GPT-4o, respectively, underscoring \revise's efficiency in sparse, question-focused reasoning.

\begin{table}[h]
\centering
\small
\caption{\footnotesize \textbf{Comparison of Training-free or Plug-and-play Methods on EgoSchema (Subset).}
Accuracy is reported in \%, and efficiency is measured as the number of frames or captions used per video.}
\vspace{-0.1in}
\begin{tabular}{lcc}
\toprule
\textbf{Method} & \textbf{Acc. (\%)} & \textbf{Frames/Captions Used} \\
\midrule
VideoAgent~\cite{wang2024videoagent} & 60.2 & 8.4 \\
VideoTree~\cite{wang2025videotree} & 66.2 & 62.4 \\
LVNet~\cite{park2026too} & 68.2 & 12 \\
LLoVi~\cite{zhang2024simple} & 57.6 & 180 \\
MC-ViT-L~\cite{balavzevic2024memory} & 62.6 & 128{+} \\
\midrule
GPT-4o~\cite{gpt4o} + \revise & 60.6 & 9.8 \\ %
\bottomrule
\vspace{-0.2in}
\label{tab:egoschema}
\end{tabular}
\end{table}

\noindent \textbf{Comparison across Baselines.}
We report a comparison with EgoSchema~\cite{mangalam2023egoschema} and NExT-QA~\cite{xiao2021next} in \Cref{tab:egoschema} and \Cref{tab:nextqa}.
On EgoSchema (subset), GPT-4o+\revise achieves 60.6\% with 9.8 frames, operating in essentially the same small-budget regime as VideoAgent~\cite{wang2024videoagent} while avoiding any captioner. Relative to selection- or caption-heavy pipelines, \revise uses far fewer visual inputs, for example, about 6$\times$ fewer than VideoTree~\cite{wang2025videotree} and over 13--18$\times$ fewer than LLoVi~\cite{zhang2024simple} and MC-ViT-L~\cite{balavzevic2024memory}, highlighting our emphasis on sparse, direct frame reasoning. 
On NExT-QA, \revise achieves 63.8\% with 8.4 frames, outperforming LVNet~\cite{park2026too} and matching SeViLA~\cite{yu2023self}, while using 3--4$\times$ fewer inputs. 
Compared with the strongest baselines, \revise trades some accuracy for simplicity and efficiency: it uses $\sim$6--7$\times$ fewer frames than VideoTree and remains in the same ultra-low-frame regime as VideoAgent without relying on caption models.
We also evaluate \revise on additional benchmarks with the same backbone models (\Cref{tab:more_benchmarks}).
\begin{table}[h]
\centering
\small
\caption{\footnotesize \textbf{Comparison of training-free or plug-and-play methods on NExT-QA.}
Results are quoted from \citet{park2026too} for fair comparison.
Accuracy is reported as average accuracy (\%), and efficiency is measured as the number of frames used per video.}
\vspace{-0.1in}
\resizebox{\columnwidth}{!}{
\begin{tabular}{lcc}
\toprule
\textbf{Method} & \textbf{Acc. (\%)} & \textbf{Frames/Captions Used} \\
\midrule
VideoTree~\cite{wang2025videotree} & 73.5 & 56 \\
VideoAgent~\cite{wang2024videoagent} & 71.3 & 8.2 \\
LLoVi~\cite{zhang2024simple} & 67.7 & 90 \\
ProViQ~\cite{choudhury2024video} & 64.6 & 60 \\
SeViLA~\cite{yu2023self} & 63.6 & 32 \\
LVNet~\cite{park2026too} & 61.1 & 12 \\
\midrule
GPT-4o~\cite{gpt4o} + \revise & 63.8 & 8.4 \\
\bottomrule
\end{tabular}
}

\label{tab:nextqa}
\vspace{-0.1in}
\end{table}

\noindent \textbf{Ablation on Frames and Turns.}
We provide detailed ablation tables in the supplementary.
\Cref{tab:ablation-best-per-turn} shows that increasing the allowed turns consistently lifts accuracy while keeping the frame budget low: the best single-turn setting reaches 38.3\% with 4.60 frames, two turns reach 39.0\% with fewer frames (3.20), and three turns reach 41.6\% at $\sim$4.0 frames.
Allowing four turns yields the best point, 42.1\% at just 2.89 frames, while average total rounds remain well below the allowed maximum ($\approx$2.3), indicating early stopping.
The accuracy--frames Pareto frontier (\Cref{fig:ablation-pareto}) is monotonic: as the controller is permitted more turns, it attains strictly better accuracy at strictly lower frame budgets.
These trends support our design that multi-round, summary-conditioned selection concentrates evidence into a few targeted frames, improving answer quality without incurring large token costs.

\noindent \textbf{Component Ablation.}
\Cref{tab:component-ablation} highlights the complementary roles of the persistent summary-as-state and its structured fields ($P/O/H/U/R$).
Removing state carryover causes large regressions in accuracy ($-$18.34\%) and nearly doubles computational cost, as the model must repeatedly reconstruct context.
Similarly, removing the structured belief/uncertainty/rationale fields leads to substantial drops and the largest runtime increase (+32.14s), indicating that explicit state propagation is critical for stable multi-round reasoning.
Ablating both yields the worst accuracy (20.24\%).
In contrast, the full model achieves the best accuracy while requiring the fewest turns and lowest latency.

\subsection{Reinforcement Fine-Tuning}
According to \cref{tab:baselines-rft},
\revise further enhances the video understanding capabilities of vision-language models (VLMs) when combined with reinforcement fine-tuning. 
In this setting, the underlying VLM is trained to make better multi-round decisions: what frames to request, how to update its summary-as-state, and when to stop, while keeping the visual encoder fixed.
\begin{table}[t]
\centering
\scriptsize
\setlength{\tabcolsep}{4pt}
\caption{\footnotesize \textbf{Comparison across baseline categories and \revise (RFT).}}
\label{tab:baselines-rft}
\vspace{-6pt}

\textbf{VideoEspresso}
\vspace{-2pt}
\begin{tabular}{lcccc}
\toprule
\textbf{Method} 
& \textbf{Acc. (\%)} 
& \textbf{Frames} 
& \textbf{Rounds} 
& \textbf{Time (s)} \\
\midrule
Direct Reasoning     & 12.6 & 8.0 & 1.00 & 1.02 \\
Plug-and-Play   & 20.1 & 5.2 & 1.86 & 1.73 \\
Supervised Format Fine-Tuning    & 21.3 & 5.0 & 1.52 & 1.67 \\
Reinforced Fine-Tuning       & \textbf{27.8} & \textbf{4.1} & \textbf{1.37} & \textbf{1.02} \\
\bottomrule
\end{tabular}

\vspace{6pt}

\textbf{NExT-QA}
\vspace{-2pt}
\begin{tabular}{lcccc}
\toprule
\textbf{Method} 
& \textbf{Acc. (\%)} 
& \textbf{Frames} 
& \textbf{Rounds} 
& \textbf{Time (s)} \\
\midrule
Direct   Reasoning       & 23.6 & 8.0 & 1.00 & 0.88 \\
Plug-and-Play       & 31.7 & 5.3 & 1.74 & 1.22 \\
Supervised Format Fine-Tuning    & 27.3 & 5.1 & 1.65 & 1.13 \\
Reinforced Fine-Tuning         & \textbf{51.3} & \textbf{3.9} & \textbf{1.32} & \textbf{0.62} \\
\bottomrule
\end{tabular}

\vspace{-0.1in}
\end{table}

\noindent \textbf{Training Setup.}
We employ the multi-round formulation in \S\ref{sec:module} and optimize the policy via GRPO~\cite{shao2024deepseekmath, guo2025deepseek}.
Specifically, we follow the same setup as RAGEN~\cite{wang2025ragen}.
We apply our verifier-guided reward (EAGER) that assigns credit based on three signals: (i) confidence gain after new frames are added, (ii) summary sufficiency, measured by whether the final answer is recoverable from the last committed \texttt{<summarize>} alone, and (iii) correct-and-early stopping that rewards answering correctly within a small turn budget. The policy is trained on the training split of each dataset using sampled trajectories (1--3 turns, depending on budget), while the summary-as-state is fully externalized and updated at every step.
We use Qwen2.5-VL-3B~\cite{bai2025qwen2_5_vl} as the backbone model due to computational cost.
Because 3B-scale VLMs have limited instruction-following capability, we first distill 8,000 multi-round conversations from GPT-4o~\cite{gpt4o} under the ``plug-and-play'' setting and perform a short format fine-tuning stage to teach the model to reliably follow the structured protocol (\texttt{<think>}, \texttt{<summarize>}, \texttt{<select>}, \texttt{<answer>}), i.e., a \texttt{<think>} trace followed by either \texttt{<summarize>} and \texttt{<select>} (request) or \texttt{<answer>} (final).
During reinforcement learning, we adopt the same configuration as \Cref{sec:exp_pnp}, using \texttt{max\_frames\_per\_round}=3 and \texttt{max\_rounds}=4.

\noindent \textbf{Baselines.}
We compare against three categories of baselines:
(i) direct video-understanding models that process all given frames in a single forward pass;
(ii) the plug-and-play setting without parameter updates; and
(iii) supervised format fine-tuning.
All methods are evaluated under a controlled frame budget (typically 3--8 frames per video) to highlight improvements in selection quality and reasoning efficiency.
\textbf{Metrics.} We report answer accuracy, the average number of frames used per example, the number of reasoning rounds, and end-to-end inference time.
All methods follow the same output constraints for fair comparison.

\noindent \textbf{Results.}
Across both datasets, \revise with reinforcement fine-tuning delivers the strongest performance while simultaneously improving efficiency. On VideoEspresso, \revise achieves 27.8\% accuracy while using only 4.1 frames and 1.37 rounds, outperforming both plug-and-play inference (20.1\%) and supervised fine-tuning (21.3\%) under the same frame constraints. Notably, \revise matches the runtime of the single-pass baseline (1.02 s) despite requiring multi-round interaction, demonstrating effective early stopping and tight summary-guided control.
On NExT-QA, the gains are even more pronounced: \revise reaches 51.3\% accuracy, surpassing plug-and-play (31.7\%) by nearly 20 percentage points, while reducing frames from 5.3 to 3.9 and rounds from 1.74 to 1.32. Reinforcement fine-tuning also cuts inference time nearly in half (0.62 s vs.\ 1.22 s). These results highlight that \revise learns to select highly discriminative frames and terminate earlier, achieving substantially better video reasoning ability than both the direct VLM and supervised fine-tuning baselines.

\section{Conclusion}
We address two challenges in long-video VLM QA: information overload and weak awareness of key evidence.
\revise is a frames-only, multi-round framework that requests a few query-relevant frames, maintains an explicit summary-as-state in \texttt{<summarize>} with fixed fields $P/O/H/U/R$, and stops early when confident.
This design concentrates evidence, keeps token usage low, and preserves cross-turn coherence without modifying the backbone VLM.
Across standard VQA benchmarks, \revise achieves competitive accuracy with single-digit frames on average, often with fewer rounds and lower prompt cost than caption-heavy or dense-frame baselines.
Reinforcement fine-tuning further improves open-source VLMs by aligning frame selection, summary quality, and stopping behavior.
It yields higher accuracy under the same frame budget.
\textbf{Limitations.} \revise still has several limitations.
(1) Backbone dependence: performance depends on the underlying VLM's visual fidelity and temporal reasoning, and weaker backbones can degrade the summary state;
(2) Interaction latency: multi-round querying introduces additional API calls; single-shot models can be faster under strict latency constraints;
(3) Frame granularity: the current system selects whole frames rather than spatial regions, which may limit efficiency on tasks requiring fine-grained localization.
Future work could explore adaptive spatial cropping, stronger vision encoders, and extending \revise to open-ended generation tasks beyond multiple-choice QA.

\clearpage
\newpage

\section{Acknowledgement}
CX would like to thank Zhenyu Pan, Jerry Yao-Chieh Hu, Jianshu Zhang, and Lining Mao for the valuable conversation, and Jiayi Wang for facilitating experimental deployments. 
ZY would like to thank Yanxun Xu for support. 
The authors would like to thank the anonymous reviewers and program chairs for constructive comments.

HL is partially supported by NIH R01LM1372201, NSF AST-2421845, Simons Foundation MPS-AI-00010513, AbbVie, Dolby, and Chan Zuckerberg Biohub Chicago Spoke Award. 
This research was supported in part through the computational resources and staff contributions provided for the Quest high-performance computing facility at Northwestern University which is jointly supported by the Office of the Provost, the Office for Research, and Northwestern University Information Technology. The content is solely the responsibility of the authors and does not necessarily represent the official views of the funding agencies.

{
    \small
    \bibliographystyle{ieeenat_fullname}
    \bibliography{main}
}

\onecolumn
\setcounter{page}{1}
\providecommand{\theHpage}{\thepage}
\renewcommand{\theHpage}{S\thepage}

\begin{center}
{\Large \textbf{\thetitle}}\\
\vspace{0.5em}
Supplementary Material \\
\vspace{1.0em}
\end{center}
\begin{table*}[!h]
    \centering
    \caption{Table of notations used in REVISE and its reinforcement fine-tuning.}
    \resizebox{\textwidth}{!}{
    \begin{tabular}{ll}
    \toprule
    Symbol & Description \\
    \midrule
    $V = \{x_i\}_{i=0}^{L-1}$     & Input video comprising $L$ frames $x_i$ \\
    $x_i$                         & The $i$-th video frame \\
    $L$                           & Number of frames in the video \\
    $p$                           & User prompt or question about the video \\
    $\pi_\theta$                  & Vision--language model (VLM) agent parameterized by $\theta$ \\
    $a$                           & Final answer generated by the agent $\pi_\theta$ \\
    $K$                           & Maximum context length (in tokens) allowed by the model \\
    $c(x_i)$                      & Visual token cost of frame $x_i$ \\
    $C(F) = \sum_{x \in F} c(x)$  & Total visual token cost of a frame set $F$ \\
    $F \subseteq V$               & Set of frames selected from $V$ for input to the VLM \\
    $|F|$                         & Number of frames in a selected set $F$ (cardinality) \\
    $T$                           & Maximum number of reasoning rounds allowed \\
    $t$                           & Round index in the multi-round reasoning process, $1 \le t \le T$ \\
    $p_t$                         & Prompt at round $t$ (with formatting and meta information) \\
    $F_t$                         & Frames shown to the agent at round $t$ \\
    $S_t = \bigcup_{j=1}^t F_j$   & Union of all frames admitted up to round $t$ (with $S_0 = \varnothing$) \\
    $z_t = (P_t, O_t, H_t, U_t, R_t)$ & Summary-as-state committed at round $t$ (fixed order $P \rightarrow O \rightarrow H \rightarrow U \rightarrow R$) \\
    $P_t$                         & Previously seen frames (summarized in natural language) \\
    $O_t$                         & Observations recorded at round $t$ in the summary \\
    $H_t$                         & Belief hypotheses/updates based on the current observations \\
    $U_t$                         & Remaining uncertainties about the video or answer \\
    $R_t$                         & Reasons for which frames to view next (or that the question is answered) \\
    $a_t$                         & Agent's action at round $t$, $a_t \in \{\mathrm{SELECT}(Q_t), \mathrm{ANSWER}(y_t)\}$ \\
    $Q_t$                         & Indices of new frames requested at round $t$ when $a_t = \mathrm{SELECT}(Q_t)$ \\
    $y_t$                         & Candidate answer produced at round $t$ when $a_t = \mathrm{ANSWER}(y_t)$ \\
    $\tau$                        & Stopping time (final round), $\tau \le T$ \\
    $s_t = (p_t, z_{t-1}, S_{t-1})$ & State in the MDP at round $t$ for reinforcement fine-tuning \\
    $r_t$                         & Reward received at round $t$ (EAGER components and formatting) \\
    $\gamma$                      & Discount factor used to compute cumulative reward in reinforcement learning \\
    $H$                           & A complete conversation trajectory (states, actions, rewards) under $\pi_\theta$ \\
    $R(H) = \sum_{t=1}^{\tau} \gamma^{t-1} r_t$ & Discounted return for trajectory $H$ in reinforcement learning \\
    $R(\cdot)$                    & Task-specific evaluation metric (e.g., accuracy) applied to the agent's final answer \\
    $|\cdot|$                     & Token length (e.g., $|p|$ or context length) \\
    \bottomrule
    \end{tabular}
    }
    \label{tab:notations}
\end{table*}

\section{Table of Notation}
\label{sec:notation_tab}
We list the table of notations of our paper in \Cref{tab:notations}.

\section{Extended Background of Reinforcement Learning in LLMs}

Reinforcement Learning (RL) enables large language models (LLMs) to improve their behavior via interaction with an environment and scalar reward feedback. Formally, given a data distribution $\mathcal{D}$ over prompts $x$ and a policy $\pi_{\theta}(y \mid x)$ parameterized by $\theta$, the RL objective maximizes the expected reward
\begin{equation}
J(\theta)
= \mathbb{E}_{x \sim \mathcal{D},\, y \sim \pi_{\theta}(\cdot \mid x)}
\bigl[ R(x,y) \bigr],
\end{equation}
where $R(x,y)$ assigns a scalar quality score to the generated output $y$ for prompt $x$.

A widely used algorithm for fine-tuning LLMs with RL is Proximal Policy Optimization (PPO)~\cite{schulman2017proximal}, which constrains policy updates by clipping the importance sampling ratio
\begin{equation}
\rho_t(\theta)
= \frac{\pi_{\theta}(y_t \mid x_t)}{\pi_{\theta_{\text{old}}}(y_t \mid x_t)}.
\end{equation}
The PPO objective maximizes a clipped surrogate with an optional KL penalty:
\begin{equation}
J_{\mathrm{PPO}}(\theta)
= \mathbb{E}_{t}\Bigl[
\min\bigl(\rho_t A_t,\; \widehat{\rho}_t A_t\bigr)
- \beta\, D_{\mathrm{KL}}\bigl[\pi_{\theta_{\text{old}}} \,\|\, \pi_{\theta}\bigr]
\Bigr],
\end{equation}
where $\widehat{\rho}_t = \operatorname{clip}(\rho_t,\, 1-\epsilon,\, 1+\epsilon)$ and $A_t$ is an advantage estimate (e.g., from a learned value network).

A common way to compute $A_t$ is Generalized Advantage Estimation (GAE)~\cite{schulman2015high}:
\begin{equation}
A_t^{\mathrm{GAE}(\gamma,\lambda)}
= \sum_{l=0}^{\infty} (\gamma \lambda)^{l}\,\delta_{t+l},
\quad
\delta_t = r_t + \gamma\,V(x_{t+1}) - V(x_t),
\end{equation}
where $\gamma$ and $\lambda$ balance bias and variance in the return estimates.

More recent work such as DeepSeekMath~\cite{shao2024deepseekmath} and DeepSeek-R1~\cite{guo2025deepseek} introduces \emph{Group Relative Policy Optimization} (GRPO), which dispenses with a learned value function and instead uses group-normalized returns. For each prompt $x$ (in our setting, a multi-round video reasoning instance), GRPO samples a group of $G$ trajectories $\{H_i\}_{i=1}^G$ from the current policy $\pi_{\theta}$. Each trajectory $H_i$ receives a scalar return $R(H_i)$, for example
\begin{equation}
R(H_i) = \sum_{t=1}^{\tau_i} \gamma^{t-1} r_{i,t},
\end{equation}
where $r_{i,t}$ is the step-wise reward and $\tau_i$ is the stopping time of trajectory $H_i$.
The returns are standardized across the group to obtain a trajectory-level advantage
\begin{equation}
\widehat{A}_i
= \frac{R(H_i) - \mathrm{mean}\bigl(\{ R(H_j) \}_{j=1}^G \bigr)}
       {\mathrm{std}\bigl(\{ R(H_j) \}_{j=1}^G \bigr)}.
\end{equation}
Let $H_i^{(n)}$ be the $n$-th token in trajectory $H_i$ and $N_i$ its length. GRPO then applies a PPO-style clipped objective at the token level, reusing the same $\widehat{A}_i$ for all tokens in $H_i$:
\begin{align}
J_{\mathrm{GRPO}}(\theta)
&= \frac{1}{G} \sum_{i=1}^{G} \frac{1}{N_i} \sum_{n=1}^{N_i}
\min\Bigl(
\rho_{i,n} \widehat{A}_i,\;
\operatorname{clip}(\rho_{i,n},\,1-\epsilon,\,1+\epsilon)\,\widehat{A}_i
\Bigr),
\\
\rho_{i,n}
&= \frac{\pi_{\theta}\bigl(H_i^{(n)} \mid H_i^{(<n)}\bigr)}
         {\pi_{\theta_{\text{old}}}\bigl(H_i^{(n)} \mid H_i^{(<n)}\bigr)}.
\end{align}
By relying on standardized returns rather than a value network, GRPO is particularly effective for multi-step reasoning tasks and has been shown to induce emergent chain-of-thought behaviors across various domains. In our main paper, we combine GRPO with the EAGER reward to fine-tune the \revise policy for efficient, summary-driven video reasoning.

\section{Detailed Algorithm}
We show the \revise main algorithm in \Cref{alg:revise}. 

\begin{algorithm*}[ht]
  \caption{\revise: Multi-Round Sparse Video Reasoning}
  \label{alg:revise}
  \begin{algorithmic}[1]

    \Require
      Video $V = \{x_i\}_{i=0}^{L-1}$;\;
      prompt $p$;\;
      max rounds $T$;\;
      max frames per round $\texttt{max\_frames\_per\_round}$;\;
      VLM agent $\pi_{\theta}$

    \State $z_0 \gets \textsc{InitSummary}()$
    \State $F_1 \gets \textsc{SampleFrames}(V,\,\texttt{max\_frames\_per\_round})$
    \State $p_1 \gets \textsc{UpdatePrompt}(p,\, z_0,\, F_1)$

    \For{$t = 1,2,\ldots,T$}

        \State $r_t \gets \pi_{\theta}(p_t,\, z_{t-1},\, F_t)$
        \State $(a_t,\, z_t) \gets \textsc{ParseResponse}(r_t)$

        \If{$a_t = \textsc{ANSWER}(y_t)$}
            \State \Return $y_t$
        \EndIf

        \State $Q_t \gets \textsc{RequestedFrames}(a_t)$
        \State $F_{t+1} \gets \textsc{ExtractFrames}(V,\, Q_t)$
        \State $p_{t+1} \gets \textsc{UpdatePrompt}(p,\, z_t,\, F_{t+1})$

    \EndFor

    \State \Return $y_t$

  \end{algorithmic}
\end{algorithm*}

\section{Experimental Details}
\label{sec:settings}

\paragraph{Computational Hardware.} 
All experiments are conducted on platforms equipped with Tesla A100 SXM GPUs (for reinforcement fine-tuning and large-scale evaluation) and Intel Xeon Silver 4214 CPUs @ 2.20GHz.

\subsection{Additional Details on Datasets}

\paragraph{VideoEspresso}~\cite{han2025videoespresso} is a large-scale chain-of-thought (CoT) video reasoning benchmark built with a core-frame selection pipeline and multimodal CoT evidence. It organizes evaluation into 14 fine-grained categories (e.g., causal, temporal, narrative), and emphasizes answering from sparse core frames rather than full video streams. In our main paper, VideoEspresso is the primary benchmark for plug-and-play experiments and reinforcement fine-tuning with \revise.

\paragraph{NExT-QA}~\cite{xiao2021next} targets causal and temporal action reasoning with both multiple-choice and open-ended QA. Videos average about 44 seconds, and questions are stratified into causal, temporal, and descriptive types. We follow the official multiple-choice split and report accuracy on the standard evaluation set. NExT-QA is used both in our plug-and-play comparisons and in reinforcement fine-tuning experiments.

\paragraph{EgoSchema}~\cite{mangalam2023egoschema} is a long-form egocentric video benchmark containing approximately 250 hours of video. Each clip is around 3 minutes and paired with a five-choice multiple-choice question, for over 5{,}000 question--answer pairs in total. Following prior work, we evaluate \revise on the official EgoSchema subset used for long-video question answering.

\subsection{Implementation Details}
For ``Plug-and-play'', we follow the same settings as the main paper.
For reinforcement fine-tuning, we highlight the important hyperparameter in \Cref{tab:hyperparameter}. 
We utilize 4$\times$80G A100 for reinforcement fine-tuning and open-source ``Plug-and-play''. 

\begin{table*}[h]
\centering
\caption{\textbf{Key Hyperparameters for PPO-based Video VLM Training.}
The listed hyperparameters are selected for their critical influence on model performance, training stability, and efficiency.}
\label{tab:hyperparameter}
\begin{tabular}{lc}
\toprule
\textbf{Hyperparameter} & \textbf{Value} \\
\midrule
Learning rate  & $1 \times 10^{-6}$ \\
KL loss coefficient & 0.001 \\
Entropy coefficient & 0 \\
PPO mini-batch size & 256 \\
batch size & 8 \\
max prompt length & 8192 \\
max response length & 512 \\
Use dynamic batch size & True \\
Total training epochs & 200 \\
\bottomrule
\end{tabular}
\end{table*}

\section{Additional Analyses}
\label{sec:additional_analyses}

\subsection{Ablation on Frames and Turns}
\begin{table}[t]
\centering
\small

\caption{\footnotesize\textbf{Ablation: best configuration per max turns.}
Accuracy is reported as in $\%$; runtime, average frames, and average total rounds are per example. The config is formed as ``max\_number\_of\_turns''\_``max\_number\_of\_frames\_per\_turn''.}
\vspace{-0.1in}
\resizebox{0.5\columnwidth}{!}{
\begin{tabular}{crrrrr}
\toprule
\textbf{Config} & \textbf{Max Turns} & \textbf{Acc. (\%)} & \textbf{Runtime (s)} & \textbf{Frames} & \textbf{Tot. Rounds} \\
\midrule
\texttt{01\_06} & 1 & 38.3 & 2.43 & 4.60 & 1.00 \\
\texttt{02\_04} & 2 & 39.0 & 3.96 & 3.20 & 1.57 \\
\texttt{03\_06} & 3 & 41.6 & 6.01 & 3.99 & 2.02 \\
\texttt{04\_04} & 4 & \textbf{42.1} & 6.26 & \textbf{2.89} & 2.28 \\
\bottomrule
\end{tabular}
}
\label{tab:ablation-best-per-turn}
\vspace{-0.2in}
\end{table}

\begin{figure}[h]
    \centering
    \includegraphics[width=0.9\columnwidth]{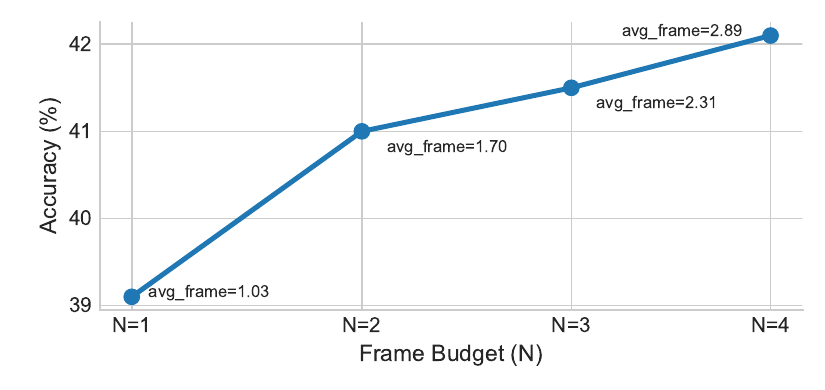}
    \vspace{-0.1in}
    \caption{\footnotesize \textbf{Accuracy--frames Pareto frontier.}
  Each point corresponds to a different frame budget $N$, and the frontier is monotone.}
    \label{fig:ablation-pareto}
\end{figure}

\noindent \Cref{tab:ablation-best-per-turn} shows that increasing the allowed turns consistently lifts accuracy while keeping the frame budget low: the best single-turn setting reaches 38.3\% with 4.60 frames, two turns reach 39.0\% with fewer frames (3.20), and three turns reach 41.6\% at $\sim$4.0 frames. Allowing four turns yields the best point, 42.1\% at just 2.89 frames, while average total rounds remain well below the allowed maximum ($\approx$ 2.3), indicating early stopping. The accuracy--frames Pareto frontier in \Cref{fig:ablation-pareto} is monotonic: as the controller is permitted more turns, it attains strictly better accuracy at strictly lower frame budgets.

\subsection{Component Ablation}
\begin{table*}[ht]
\centering
\setlength{\tabcolsep}{4pt}
\caption{\footnotesize \textbf{Component ablation (plug-and-play).}
We ablate (i) state carryover and/or (ii) the structured summary fields ($P/O/H/U/R$) against the full configuration.
Accuracy is average answer accuracy; \#Turns and Time are per example; $\uparrow$/$\downarrow$ indicate desired direction.
For ablated rows we also report deltas vs.\ \textbf{Full} in parentheses.}
\vspace{-0.1in}

\resizebox{0.8\textwidth}{!}{
\begin{tabular}{lrrrr}
\toprule
\textbf{Variant} & \textbf{Acc. (\%) $\uparrow$} & \textbf{Avg Turns $\downarrow$} & \textbf{Time (s) $\downarrow$} & \textbf{Early Stop (\%)} \\
\midrule
\textbf{Full (structured summary-as-state)} & \textbf{41.48} & \textbf{2.79} & \textbf{22.71} & 99.9 \\
\midrule
No state carryover & 23.14 \,(\,$-18.34$\,) & 3.73 \,(\,$+0.94$\,) & 50.75 \,(\,$+28.04$\,) & 100.0 \\
No structured $P/O/H/U/R$ fields & 24.27 \,(\,$-17.21$\,) & 3.91 \,(\,$+1.12$\,) & 54.85 \,(\,$+32.14$\,) & 100.0 \\
No state carryover \& No structured fields & 20.24 \,(\,$-21.24$\,) & 3.08 \,(\,$+0.29$\,) & 45.81 \,(\,$+23.10$\,) & 99.8 \\
\bottomrule
\end{tabular}
}

\label{tab:component-ablation}
\vspace{-0.1in}
\end{table*}

\noindent \Cref{tab:component-ablation} highlights the complementary roles of the persistent summary-as-state and its structured fields ($P/O/H/U/R$). Removing state carryover causes large regressions in accuracy ($-$18.34\%) and nearly doubles computational cost. Similarly, removing the structured fields leads to substantial drops and the largest runtime increase (+32.14s). Ablating both yields the worst accuracy (20.24\%). The full model achieves the best accuracy while requiring the fewest turns and lowest latency.

\subsection{Turn-budget Allocation}

\begin{table}[h]
\centering
\footnotesize
\caption{\footnotesize \textbf{Turn-budget allocation ablation.}
We compare different allocations of \texttt{max\_rounds} and \texttt{max\_frames\_per\_round} under a comparable total frame budget. Balanced allocations (e.g., $5{\times}4$ and $4{\times}5$) outperform skewed ones (e.g., $10{\times}2$ and $2{\times}10$).}
\label{tab:budget_allocation_ablation}
\begin{tabular}{c|c|c|c|c}
\toprule
Max Rounds & Max Frames / Round & Accuracy (\%) & Avg Rounds & Avg Frames Used \\
\midrule
10 & 2 & 63.3 & 9.93 & 19.70 \\
2 & 10 & 64.7 & 1.98 & 17.42 \\
4 & 5 & 66.8 & 3.93 & 16.89 \\
\cellcolor{LightCyan} 5  & \cellcolor{LightCyan} 4  & \cellcolor{LightCyan}  \textbf{66.9} & 4.93 & 17.17 \\
\midrule
\cellcolor{LightGray} 7  & \cellcolor{LightGray} 4  & \cellcolor{LightGray} 67.4 & 6.83 & 23.09  \\
\cellcolor{LightGray} 9  & \cellcolor{LightGray} 4  & \cellcolor{LightGray} 66.8 & 8.84 & 29.25 \\
\bottomrule
\end{tabular}
\end{table}

\noindent Under a fixed total frame budget (20), balanced allocations (e.g., $5{\times}4$ and $4{\times}5$) consistently outperform skewed ones (e.g., $10{\times}2$ and $2{\times}10$), suggesting that distributing a limited frame budget over more interactive rounds can be beneficial. Increasing the allowed rounds budget can further help when the frame budget is correspondingly increased (e.g., $7{\times}4$ vs.\ $5{\times}4$), though gains diminish with very large budgets.

\subsection{Caption Effect Analysis}
\begin{table}[h]
  \centering
  \caption{\footnotesize \textbf{Caption effect analysis.}
We compare \revise with and without captions on NExT-QA question types (Descriptive/Temporal/Causal). \textit{Tok/sample} reports total text tokens used (caption tokens shown as ``+caption'').}
  \label{tab:caption_effect_ablation}
  \resizebox{0.7\linewidth}{!}{\begin{tabular}{l|cccc|c|cccc|c}
    \toprule
    \multirow{3}{*}{\shortstack{Setting (NExT-QA).\\ Tok: Total token used}}
      & \multicolumn{5}{c|}{Qwen7B}
      & \multicolumn{5}{c}{GPT-4o} \\
    \cmidrule(lr){2-6}\cmidrule(lr){7-11}
      & \multicolumn{4}{c|}{Accuracy (\%)} 
      & \multirow{2}{*}{Tok/sample}
      & \multicolumn{4}{c|}{Accuracy (\%)} 
      & \multirow{2}{*}{Tok/sample} \\
    \cmidrule(lr){2-5}\cmidrule(lr){7-10}
      & D & T & C & Avg
      & 
      & D & T & C & Avg
      &  \\
    \midrule
    VideoAgent
      & \cellcolor{LightGray}75.4 & \cellcolor{LightGray}\textbf{60.2} & \cellcolor{LightGray}59.8 & \cellcolor{LightGray}\textbf{65.1}
      &4810+3749
      & \cellcolor{LightGray}79.9 & \cellcolor{LightGray}\textbf{69.6} & \cellcolor{LightGray}71.5 & \cellcolor{LightGray}73.7
      & 4025+3749 \\
    \revise (frames)
      & \cellcolor{LightCyan} \textbf{75.7} & \cellcolor{LightCyan}53.9 & \cellcolor{LightCyan}\textbf{63.7} & \cellcolor{LightCyan}64.4
      & \textbf{6647}
      & \cellcolor{LightCyan}78.0 & \cellcolor{LightCyan}67.8 & \cellcolor{LightCyan}\textbf{76.6} & \cellcolor{LightCyan}74.1
      & \textbf{3089} \\
    \revise (frames + caption)
      & 75.5 & 56.7 & 61.3 & 64.5 & 12421+3749
      & \textbf{81.8} & 68.7 & 75.8 & \textbf{75.4} & 4433+3749 \\
    \revise (caption-only)
      & 69.5 & 50.7 & 55.8 & 58.7 & 6441+3749
      & 76.7 & 61.7 & 69.7 & 69.4 & 2654+3749 \\
    \bottomrule
  \end{tabular}}
\end{table}

\noindent Captions can help when selected frames miss globally relevant evidence (e.g., counting), but can hurt when decisions rely on fine-grained visual cues (e.g., causal ``why'' questions where captions emphasize irrelevant context). Overall, captions tend to help temporal questions more than causal questions under the same frame constraints.

\subsection{Summary-as-State Drift Analysis}
\noindent We analyze errors to understand whether failures are dominated by missing visual evidence or by drift in the persistent summary state. Empirically, early-stopping accounts for a negligible fraction of errors ($\leq 0.2\%$). Most errors are attributed to missing key evidence (70--82\%), while summary-state drift contributes a smaller fraction (15--20\%), mainly on temporal questions. Increasing the maximum number of rounds ($5\rightarrow 9$) does not significantly change accuracy, but can increase formatting failures, indicating diminishing returns rather than premature stopping.

\subsection{Reward Robustness}
\begin{table}[h]
\centering
\footnotesize
\caption{\footnotesize \textbf{EAGER reward ablation.}
We vary reward weights and early-stop parameters ($\beta, T_{\text{stop}}$) to evaluate robustness.}
\label{tab:reward_ablation}{
\begin{tabular}{l|c|c|c}
\toprule
Setting
& $\lambda_1,\lambda_2,\lambda_3, \beta, T_{\text{stop}}$
& Acc (\%)
& Avg Rounds \\
\midrule
\textbf{Base (no RL)}
& n/a
& 31.7
& 1.74 \\
\midrule
\multicolumn{4}{l}{\textit{Reward weight ablation (fixed $\beta{=}1.0,\,T_{\text{stop}}{=}2)$}} \\
EAGER
& $\lambda_1{=}1.0,\ \lambda_2{=}1.0,\ \lambda_3{=}0.5$
& 44.4
& 2.778 \\
EAGER
& $\lambda_1{=}0.0,\ \lambda_2{=}1.0,\ \lambda_3{=}0.5$
& \textbf{46.4}
& 2.492 \\
EAGER
& $\lambda_1{=}1.0,\ \lambda_2{=}0.0,\ \lambda_3{=}0.5$
& 44.8
& 2.620 \\
EAGER
& $\lambda_1{=}1.0,\ \lambda_2{=}1.0,\ \lambda_3{=}0.0$
& 39.9
& 3.032 \\
\midrule
\multicolumn{4}{l}{\textit{Early-stop design ablation (fixed $\lambda_1{=}0,\lambda_2{=}1,\lambda_3{=}0.5$)}} \\
EAGER
& $\beta{=}0.0,\,T_{\text{stop}}{=}1$
& 47.1
& 3.115 \\
EAGER
& $\beta{=}0.0,\,T_{\text{stop}}{=}3$
& 49.8
& 3.410 \\
EAGER
& $\beta{=}1.0,\,T_{\text{stop}}{=}2$
& \textbf{49.5}
& 3.155 \\
\midrule
Paper Setting & $\lambda_1{=}1.0, \lambda_2{=}1.0, \lambda_3{=}0.5, \beta{=}1.0, \,T_{\text{stop}}{=}2$&\textbf{51.3} & 1.320 \\
\bottomrule
\end{tabular}
}
\end{table}

\noindent \Cref{tab:reward_ablation} shows that EAGER is robust to reward hyperparameters, and that the early-stop component helps improve both accuracy and efficiency.

\subsection{More Benchmarks}
\begin{table}[h]
  \centering
  \small
  \caption{\footnotesize \textbf{More benchmarks.}
  Comparison on Video-MME and LVBench under the same backbone (LLaVA-OV-7B). We report accuracy and average frames used.}
  \label{tab:more_benchmarks}
 \begin{tabular}{lcc}
    \toprule
    Method (backbone: LLAVA-OV-7B)
    & Video-MME (Acc / Frames)
    & LVBench (Acc / Frames) \\
    \midrule
    Adaptive Keyframes (CVPR 25)
    & 58.4 / 32
    & 59.3 / 32 \\
    MDP3 (ICCV 25)
    & 59.6 / 8
    & 59.0 / 8 \\
    \revise
    & \textbf{60.7} / 7.38
    & \textbf{61.5} / 5.62 \\
    \bottomrule
  \end{tabular}
\end{table}

\noindent We additionally evaluate \revise on Video-MME and LVBench under the same backbone and report competitive accuracy with substantially fewer frames.

\section{Broader Impacts}
As a new methodology for improving video understanding, we do not identify any direct negative societal impacts such as disinformation, unfair decision-making, privacy violations, or security risks. 
Instead, \revise facilitates more efficient and accurate analysis of existing videos, benefiting downstream research and potentially enabling applications in education, accessibility, and scientific video analysis.

\end{document}